\documentclass[10pt,twocolumn,letterpaper]{article}

\usepackage{iccv}
\usepackage{times}
\usepackage{epsfig}
\usepackage{graphicx}
\usepackage{amsmath}
\usepackage{amssymb}

\usepackage{xspace} %
\usepackage{graphicx}
\usepackage{subcaption}
\usepackage{multirow}
\usepackage{lineno}
\usepackage{booktabs}
\usepackage{bbding}        %
\usepackage[table]{xcolor} %
\usepackage{colortbl}
\usepackage{amsmath}
\usepackage{pdfrender}
\usepackage{hhline}
\usepackage[numbers]{natbib}
\definecolor{mygray}{rgb}{0.7, 0.7, 1.0}
\definecolor{mygray2}{gray}{0.9}
\definecolor{myblue}{rgb}{0.8, 0.8, 1.0}

\usepackage[pagebackref=true,breaklinks=true,letterpaper=true,colorlinks,bookmarks=false]{hyperref}

\iccvfinalcopy %

\makeatletter\renewcommand\paragraph{\@startsection{paragraph}{4}{\z@}
	{.25em \@plus1ex \@minus.2ex}{-.5em}{\normalfont\normalsize\bfseries}}\makeatother
\ificcvfinal\fi

\begin{document}

\title{\textsf{OpenVision}\,\includegraphics[width=.6cm]{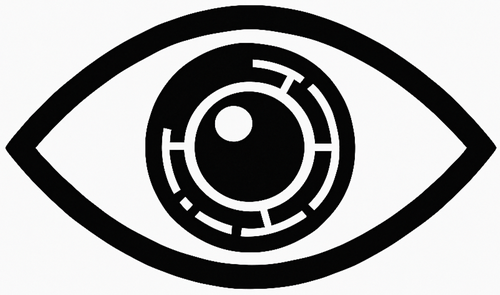}: A \textit{Fully-Open}, \textit{Cost-Effective} Family of Advanced \\Vision Encoders for Multimodal Learning}

\author{%
  Xianhang Li\thanks{Equal contribution.} \, \, 
  Yanqing Liu\footnotemark[1] \, \,
  Haoqin Tu \, \,
  Hongru Zhu \, \,
  Cihang Xie \vspace{.3em}\\ 
 University of California, Santa Cruz \vspace{.5em}
  \\
  \small
  \hspace{3em} \includegraphics[height=1.1em]{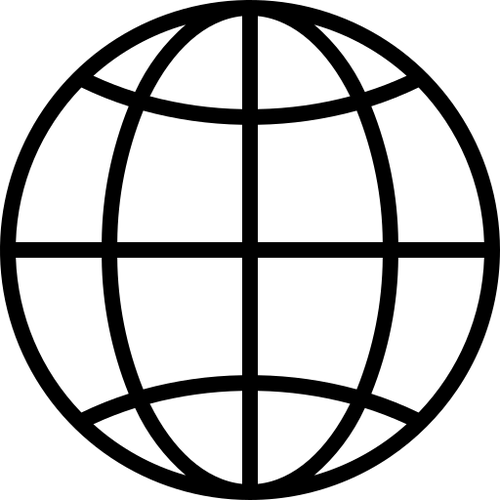} \textbf{Project Page}: \url{https://ucsc-vlaa.github.io/OpenVision} \\
  \small
  \hspace{3em} \includegraphics[height=1.2em]{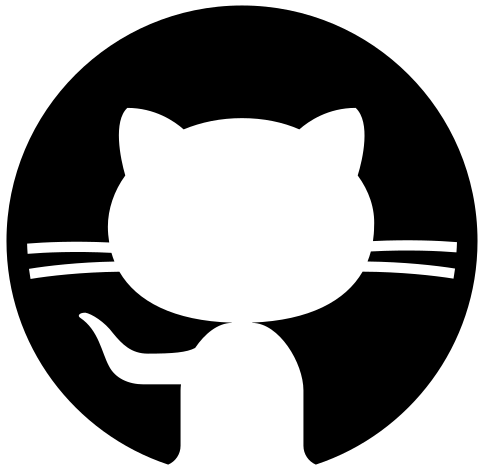} \textbf{Model Training}: \url{https://github.com/UCSC-VLAA/OpenVision} \\
  \small
  \hspace{3em} \includegraphics[height=1.2em]{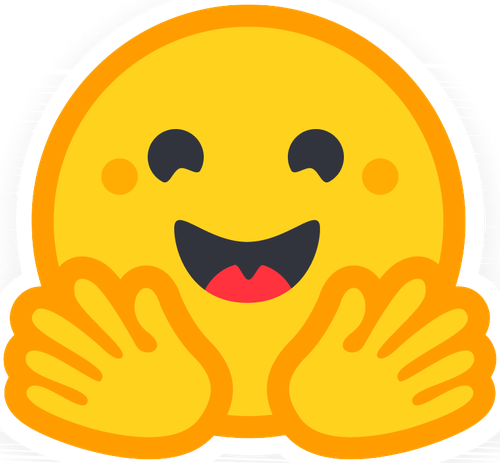} \textbf{Model Zoo}: \href{https://huggingface.co/collections/UCSC-VLAA/openvision-681a4c27ee1f66411b4ae919}{click me}
}

\maketitle
\ificcvfinal\thispagestyle{empty}\fi

\begin{abstract}
OpenAI's CLIP, released in early 2021, have long been the go-to choice of vision encoder for building multimodal foundation models. Although recent alternatives such as SigLIP have begun to challenge this status quo, to our knowledge none are fully open: their training data remains proprietary and/or their training recipes are not released.
This paper fills this gap with \textsf{OpenVision}, a \textbf{fully-open}, \textbf{cost-effective} family of vision encoders that match or surpass the performance of OpenAI’s CLIP when integrated into multimodal frameworks like LLaVA. \textsf{OpenVision} builds on existing works---\eg, \emph{CLIPS} for training framework and \emph{Recap-DataComp-1B} for training data---while revealing multiple key insights in enhancing encoder quality and showcasing practical benefits in advancing multimodal models.  By releasing vision encoders spanning from 5.9M to 632.1M parameters, \textsf{OpenVision} offers practitioners a flexible trade-off between capacity and efficiency in building multimodal models: larger models deliver enhanced multimodal performance, while smaller versions enable lightweight, edge-ready multimodal deployments.
\end{abstract}

\begin{figure}[t!]
    \centering
    \vspace{-.8em}
    \includegraphics[width=\linewidth]{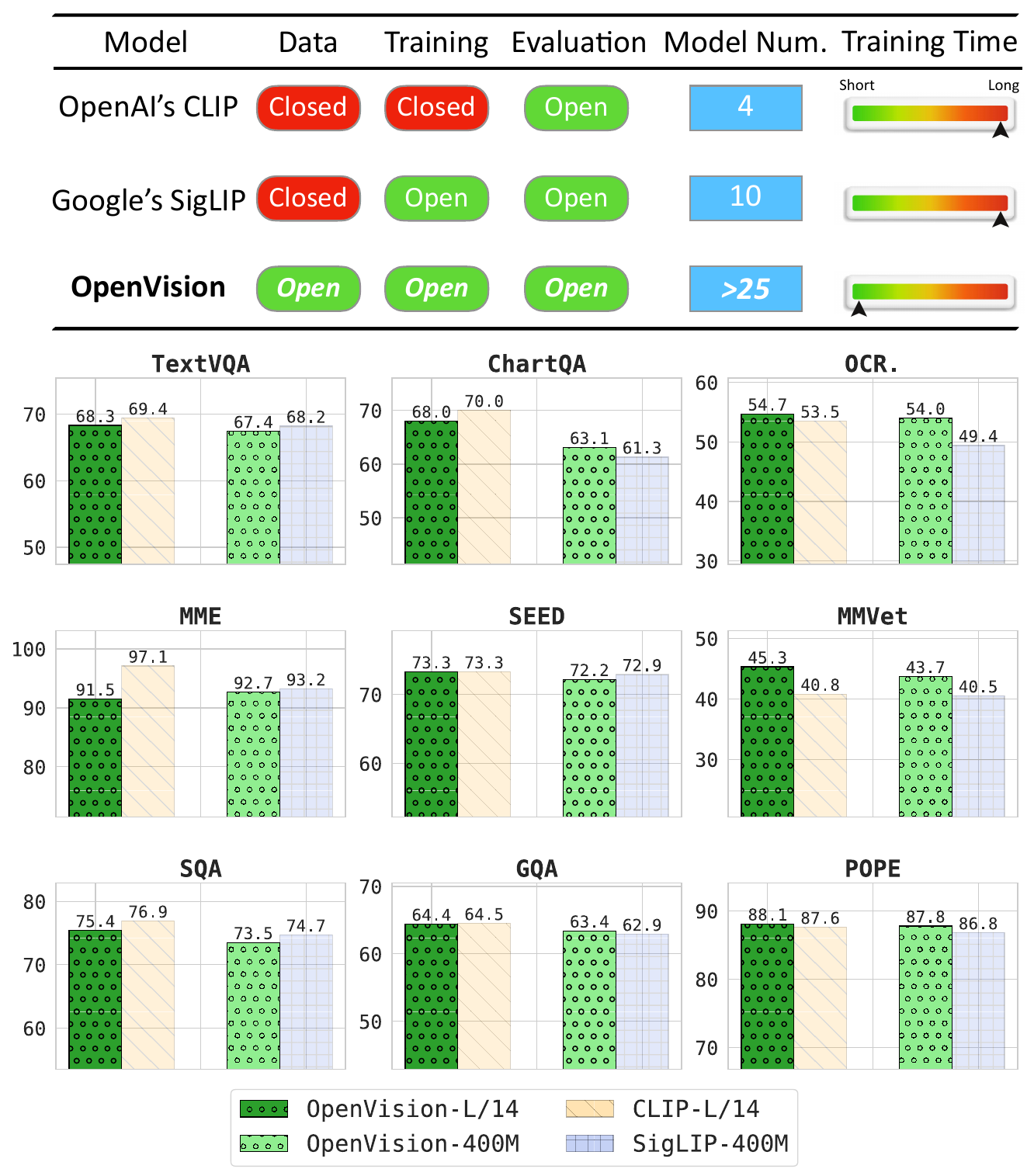}
    \vspace{-1.8em}
    \caption{The \textit{top} table compares our \textsf{OpenVision} series to OpenAI's CLIP and Google's SigLIP. The \textit{bottom} figure showcases that \textsf{OpenVision} attain competitive or even superior multimodal performance than OpenAI's CLIP and Google's SigLIP.}
    \label{fig:openvision_teaser}
    \vspace{-1.1em}
\end{figure}

\section{Introduction}

Recent advances in multimodal foundation models rely almost exclusively on the same visual backbone: OpenAI’s CLIP encoders~\cite{radford2021clip}.
From early open-source efforts such as LLaVA~\cite{llava} and Mini-GPT-4~\cite{zhu2023minigpt}, to the most recent advanced models such as Falcon2 VLM~\cite{malartic2024falcon2} and Eagle~\cite{shieagle}, OpenAI's CLIP-L/336 has consistently been the default choice, even as the language components have evolved rapidly. 
This dependence, however, imposes several issues. First, OpenAI CLIP's training data and detailed framework remain undisclosed, limiting transparency and reproducibility. Moreover, OpenAI's CLIP is available only in two parameter scales---Base and Large---hindering both the deployment of lightweight models on edge devices and the exploration of higher-capacity encoders for complex tasks. Finally, OpenAI's CLIP suffers from documented weaknesses, including poor spatial-relation understanding and object-counting hallucinations~\cite{tong2024eyes,tong2025cambrian,tu2023many}. These shortcomings call for a vision encoder whose architecture, data, and training recipe are fully open.

In response, the open-source community has mounted a concerted effort to replicate and surpass OpenAI's CLIP, notably through (1) fully open CLIP training frameworks~\cite{openclip}, (2) billion-scale open datasets such as Laion~\cite{schuhmann2022laion}, DataComp~\cite{gadre2023datacomp}, and DFN~\cite{dfn}, and (3) improved training methodologies~\cite{li2023clipa,li2023clipav2,li2023reclip,zhai2023sigmoid}.
Yet a crucial gap persists: no fully open, from-scratch vision encoder of comparable capacity and resolution consistently matches—or surpasses—OpenAI’s CLIP when used as the visual backbone of multimodal foundation models. For example, popular OpenCLIP \cite{openclip} checkpoints achieve superior zero-shot performance, but they fall markedly short on multimodal benchmarks such as MME \cite{fu2023mme}, ChartQA \cite{masry-etal-2022-chartqa} and TextVQA \cite{singh2019towards} (see Tables \ref{Tab:llava1.5_main1} and \ref{Tab:llava-next_main1}).

In this work, we address this gap with \textsf{OpenVision}, a fully-open, cost-effective family of vision encoders that excel in multimodal learning scenarios (Figure~\ref{fig:openvision_teaser}\footnote{Note that we normalize OCR and MME scores to the range of 0 to 100 following previous research~\cite{fini2024multimodal}.}). 
\textsf{OpenVision} builds on two recent advances: (i) Recap-DataComp-1B \cite{li2024if}, which re-captions the entire DataComp-1B corpus~\cite{gadre2023datacomp} using a LLaVA model powered by Llama-3 ~\cite{llama3}; and (ii) CLIPS \cite{liu2024clips}, an enhanced CLIP training pipeline that incorporates synthetic captions. Leveraging these resources, we conduct a systematic analysis to identify key design elements that improve overall training efficiency and enhance the quality of vision encoders, as well as showcasing their practical benefits in the development of different multimodal models.

Extensive experiments show that \textsf{OpenVision} matches---and sometimes exceeds---OpenAI’s CLIP across a suite of multimodal evaluations when used as the visual backbone of multimodal models such as LLaVA-1.5 and Open-LLaVA-Next. To accommodate diverse deployment needs, we release more than 25 checkpoints ranging from 5.9 million to 632.1 million parameters, enabling smooth accuracy–efficiency trade-offs from edge devices to high-capacity servers. By openly releasing datasets, training recipes, and checkpoints,  we hope \textsf{OpenVision} can set a new standard for transparency and flexibility, enabling the community to push multimodal research beyond the constraints of proprietary encoders.

\section{\textsf{OpenVision} Training and Evaluation}
\label{Sec:train_eval}
This section outlines the pipeline for building and assessing the \textsf{OpenVision} family. 
We provide details about the vision encoder pre-training, multimodal large language model (MLLM) instruction tuning, and MLLM evaluation.

\begin{table*}[t!]
    \centering
    \caption{Comparison of \textsf{OpenVision} encoders with existing CLIP variants on CLIP benchmarks and multimodal downstream tasks under the LLaVA-1.5 framework. Cls./Retr.: zero-shot classification accuracy on ImageNet or image and text retrieval on MSCOCO. \textsf{OpenVision} outperforms OpenAI-CLIP significantly across multiple settings.}
    \label{Tab:llava1.5_main1}
    \vspace{-.6em}
    \resizebox{\linewidth}{!}{
    \begin{tabular}{c|c|c|c|c|c|c|c|c|c|c|c|c|c}
    \toprule
    \multirow{2}{*}{\textbf{Method}} &
    \multirow{2}{*}{\textbf{Vision Encoder}} &
    \multirow{2}{*}{\textbf{\# Res.}} &
    \multicolumn{2}{c|}{\textbf{CLIP-Bench}} &
    \multirow{2}{*}{\textbf{Text VQA}} &
    \multirow{2}{*}{\textbf{Chart QA}} &
    \multirow{2}{*}{\textbf{OCR.}} &
    \multirow{2}{*}{\textbf{MME}} &
    \multirow{2}{*}{\textbf{SEED}} &
    \multirow{2}{*}{\textbf{MMVet}} &
    \multirow{2}{*}{\textbf{SQA}} &
    \multirow{2}{*}{\textbf{GQA}}  &
    \multirow{2}{*}{\textbf{POPE}} \\ 
    \cline{4-5}
    & & & \textbf{Cls.} & \textbf{Retr.} & & & & & & & & & \\
    \midrule
     OpenAI-CLIP~\citep{radford2021clip} & B/16 &224
    &68.3
    &33.1/52.4
    &53.1
    &11.9
    &153
    &1444/325
    &63.7
    &28.3
    &72.5
    &59.9
    &83.4\\
    SigLIP~\citep{zhai2023sigmoid} &B/16 &224  
    &76.0
    &47.8/65.7
    &53.3
    &12.2
    &238
    &1421/318
    &65.5
    &31.3
    &73.8
    &60.3
    &84.2\\
    \rowcolor{cyan!10}\textsf{OpenVision} & \cellcolor{cyan!10}B/16 & \cellcolor{cyan!10}224 
    &\cellcolor{cyan!10}73.9
    &\cellcolor{cyan!10}51.1/71.6
    &54.1
    &11.8
    &262
    &1496/293
    &68.2
    &30.9
    &74.4
    &61.6
    &86.6\\
    \midrule
    SigLIP~\citep{zhai2023sigmoid} &B/16 &384  
    &78.5
    &49.9/67.7
    &57.3
    &13.9
    &285
    &1411/266
    &67.7
    &33.6
    &73.2
    &62.0
    &86.0\\
    \rowcolor{cyan!10}\textsf{OpenVision} & \cellcolor{cyan!10}B/16 & \cellcolor{cyan!10}384 
    &\cellcolor{cyan!10}74.5
    &\cellcolor{cyan!10}52.0/72.3 
    &57.9
    &14.5
    &293
    &1432/333
    &69.8
    &33.2
    &73.5
    &62.8
    &87.8\\
    \midrule
    OpenAI-CLIP~\citep{radford2021clip} & L/14 & 224  
    &75.5  &36.5/56.3 
    &56.1
    &13.2
    &177
    &1443/306
    &66.0
    &32.8
    &73.4
    &60.8
    &85.0\\
    LAION-2B-CLIP~\citep{openclip} & L/14 & 224  &75.3 &46.5/63.4
    &54.2
    &12.8
    &165
    &1434/298
    &65.5
    &31.4
    &76.0
    &59.0
    &84.5\\
    DataComp-1B-CLIP~\citep{gadre2023datacomp} & L/14 & 224  &79.2&45.7/63.3
    &53.0
    &12.3
    &131
    &1382/312
    &62.4
    &28.9
    &74.2
    &57.8
    &83.0\\
    DFN-2B-CLIP~\citep{dfn} & L/14 & 224&81.4 &48.6/65.6
    &53.2
    &12.4
    &246
    &1447/306
    &65.6
    &29.4
    &76.3
    &59.1
    &85.0\\
    MetaCLIP-5B~\citep{metaclip} & L/14 & 224&79.2 &47.1/64.4
    &55.6
    &12.8
    &313
    &1552/315
    &67.4
    &34.6
    &78.0
    &61.3
    &85.4\\
    \rowcolor{cyan!10} \textsf{OpenVision} & \cellcolor{cyan!10}L/14 & \cellcolor{cyan!10}224 &\cellcolor{cyan!10}78.4  & \cellcolor{cyan!10}55.3/75.2 
    &57.7
    &13.9
    &315
    &1487/317
    &69.5
    &35.2
    &73.6
    &62.9
    &86.4\\
    \midrule
     OpenAI-CLIP~\citep{radford2021clip} & L/14 &336 
     &76.6 &37.1/57.9
     &59.1
    &13.8
    &201
    &1475/288
    &67.5
    &35.2
    &73.1
    &61.1
    &85.7
    \\
    \rowcolor{cyan!10} \textsf{OpenVision} & \cellcolor{cyan!10}L/14 & \cellcolor{cyan!10}336 &78.8 &55.9/75.2 &61.2
    &15.7
    &339
    &1525/315
    &70.5
    &36.2
    &75.1
    &63.7
    &87.2
\\
     \midrule
     SigLIP~\citep{zhai2023sigmoid} &SoViT-400M/14 &384
     &83.2 &52.0/70.2
     &62.6
&14.5
&338
&1481/347
&69.4
&35.1
&76.7
&63.3
&87.0 \\
   \rowcolor{cyan!10} \textsf{OpenVision} & SoViT-400M/14 &384 & 79.9  & 57.6/77.5 
    &62.4
&16.1
&357
&1493/320
&70.4
&35.3
&72.4
&63.8
&88.0\\
    \bottomrule   
    \end{tabular}}
    \vspace{-.5em}
\end{table*}

\subsection{Vision Encoder Pre-training}
\label{Sec:vision_encoder_pretrain}
Recent studies have revealed multiple key aspects in advancing MLLMs, including model architecture and training strategies~\cite{deitke2024molmo,chen2024expanding,tong2025cambrian}, yet the discussion about its vision encoder training remains lacking.
Our objectives are therefore two-fold:
(i) to publish a fully reproducible ``from-scratch'' recipe for training strong vision encoders, and
(ii) to isolate the design choices that matter most once these encoders are paired with an LLM.

We leverage CLIPS~\cite{liu2024clips}---a recent variant of CLIP---as our building foundation.
CLIPS employs the standard two-tower architecture with a contrastive objective, but extends it with a \emph{multi-positive} loss that treats both the original and synthetic captions of an image as positives.  A lightweight text decoder is trained jointly to generate new captions, further enriching the training signal. 
While CLIPS attains state-of-the-art zero-shot retrieval performance, its suitability as an MLLM perception module remains underexplored---a gap we fill in this work. Additionally, following CLIPS, we use Recap-DataComp-1B~\cite{li2024if}, a re-captioned version of the billion-scale DataComp corpus~\cite{gadre2023datacomp}~\cite{llama3}, for training.  Both CLIPS codebase~\footnote{https://ucsc-vlaa.github.io/CLIPS/} and Recap-DataComp-1B dataset~\footnote{https://www.haqtu.me/Recap-Datacomp-1B/} are fully open-sourced.

\paragraph{Training Stages and Resolution.} 
Following the efficient training curriculum of CLIPA~\cite{li2023clipa,li2023clipav2},  we pre-train every encoder in three successive resolution stages. Specifically, the Large, SoViT-400M, and Huge variants are trained at $84\times84$, $224\times224$, and finally $336\times336$ or $384\times384$.  
Smaller models such as Tiny, Small, and Base start at a larger resolution of $160\times160$, and then continues with $224\times224$, and $336\times336$ or $384\times384$.  
This staged approach substantially improves efficiency and naturally yields model variants capable of handling different input resolutions. After pre-training, we discard the text tower and decoder, retaining only the vision backbone.

\paragraph{Training Details.} Across three stages, each models processes 12.8B, 1.024B, and 256M image–text pairs, respectively. The global batch sizes are 32K, 16K, and 8K, with cosine-decayed base learning rates of  $8\times10^{-6}$, $4\times10^{-7}$, and $1\times10^{-7}$.  
The text encoder uses 80 input tokens, and the text decoder generates 128 tokens, consistent with CLIPS~\cite{liu2024clips}.  
For experiments involving different patch sizes, we only modify the patch size to 8; fixed sine-cosine positional embeddings allow adaptation to varying sequence lengths.

\begin{table*}[t!]
    \centering
    \caption{Comparison of \textsf{OpenVision} encoders with existing CLIP variants on CLIP benchmarks and multimodal downstream tasks under the Open-LLaVA-Next framework. Cls./Retr.: zero-shot classification accuracy on ImageNet or image and text retrieval on MSCOCO. \textsf{OpenVision} achieves comparable or even better performance than existing models.}
    \label{Tab:llava-next_main1}
    \vspace{-.6em}
    \resizebox{\linewidth}{!}{
    \begin{tabular}{c|c|c|c|c|c|c|c|c|c|c|c|c|c}
    \toprule
    \multirow{2}{*}{\textbf{Method}} &
    \multirow{2}{*}{\textbf{vision Encoder}} &
    \multirow{2}{*}{\textbf{\# Res.}} &
    \multicolumn{2}{c|}{\textbf{CLIP-Bench}} &
    \multirow{2}{*}{\textbf{Text VQA}} &
    \multirow{2}{*}{\textbf{Chart QA}} &
    \multirow{2}{*}{\textbf{OCR.}} &
    \multirow{2}{*}{\textbf{MME}} &
    \multirow{2}{*}{\textbf{SEED}} &
    \multirow{2}{*}{\textbf{MMVet}} &
    \multirow{2}{*}{\textbf{SQA}} &
    \multirow{2}{*}{\textbf{GQA}}  &
    \multirow{2}{*}{\textbf{POPE}} \\ 
    \cline{4-5}
    & & & \textbf{Cls.} & \textbf{Retr.} & & & & & & & & & \\
    \midrule
     OpenAI-CLIP~\citep{radford2021clip} & B/16 &224
    &68.3
    &33.1/52.4
    &58.7
    &57.5
    &379
    &1497/321
    &70.0
    &38.6
    &74.0
    &62.7
    &86.6\\
    SigLIP~\citep{zhai2023sigmoid} &B/16 &224  
    &76.0
    &47.8/65.7
    &58.4
    &53.6
    &377
    &1430/332
    &69.5
    &33.6
    &75.9
    &62.4
    &85.8\\
    \rowcolor{cyan!10}\textsf{OpenVision} & \cellcolor{cyan!10}B/16 & \cellcolor{cyan!10}224 
    &\cellcolor{cyan!10}73.9
    &\cellcolor{cyan!10}51.1/71.6 
    &\cellcolor{cyan!10}{60.7}  
    &\cellcolor{cyan!10}{59.2}
    &\cellcolor{cyan!10}{405}  
    &\cellcolor{cyan!10}{1520/336}
    &\cellcolor{cyan!10}71.8  
    &\cellcolor{cyan!10}{38.8}  
    &\cellcolor{cyan!10}{73.1} 
    &\cellcolor{cyan!10}{63.1}  
    &\cellcolor{cyan!10}86.4 \\
    \midrule
    SigLIP~\citep{zhai2023sigmoid} &B/16 &384  
    &78.5
    &49.9/67.7
    &64.2
    &63.3
    &476
    &1540/326
    &71.3
    &38.7
    &69.0
    &62.6
    &87.6\\
    \rowcolor{cyan!10}\textsf{OpenVision} & \cellcolor{cyan!10}B/16 & \cellcolor{cyan!10}384 
    &\cellcolor{cyan!10}74.5
    &\cellcolor{cyan!10}52.0/72.3  
    &\cellcolor{cyan!10}{66.3}
    &\cellcolor{cyan!10}{67.4}
    &\cellcolor{cyan!10}{499}
    &\cellcolor{cyan!10}1501/330
    &\cellcolor{cyan!10}{72.9}
    &\cellcolor{cyan!10}{40.6}
    &\cellcolor{cyan!10}{69.8}
    &\cellcolor{cyan!10}{64.0}
    &\cellcolor{cyan!10}{87.7}\\
    \midrule
    OpenAI-CLIP~\citep{radford2021clip} & L/14 & 224  &75.5  &36.5/56.3  & 62.8 & 60.7 & 459 & 1600/334 & 70.6 & 41.5 & 75.0 & 62.8  & 86.9 \\
    LAION-2B-CLIP~\citep{openclip} & L/14 & 224  &75.3 &46.5/63.4  &59.4 &50.8 &396 &1533/323 &70.0 &36.2 &72.9 &62.7  &86.4 \\
    DataComp-1B-CLIP ~\citep{gadre2023datacomp} & L/14 & 224  &79.2&45.7/63.3  &58.1 &48.5 &373 &1524/348 &70.2 &37.2 &75.6 &62.3  &86.2\\
    DFN-2B-CLIP~\citep{dfn} & L/14 & 224&81.4 &48.6/65.6 &57.0 &42.7 &303 &1486/328 &68.3 &34.5 &70.6 &61.7 &86.0 \\
    MetaCLIP-5B~\citep{metaclip} & L/14 & 224& 79.2&47.1/64.4 & 63.0 & 62.9 & 493& 1590/335& 72.3& 41.8& 77.1& 64.0& 86.8\\
    \rowcolor{cyan!10} \textsf{OpenVision} & \cellcolor{cyan!10}L/14 & \cellcolor{cyan!10}224 
    &\cellcolor{cyan!10}78.4  & \cellcolor{cyan!10}55.3/75.2  &\cellcolor{cyan!10}{65.7} & \cellcolor{cyan!10}{61.5}  & \cellcolor{cyan!10} {503}  & \cellcolor{cyan!10}1567/332  &\cellcolor{cyan!10}{73.1} &\cellcolor{cyan!10}{41.4} 
    &\cellcolor{cyan!10}{73.1}
    &\cellcolor{cyan!10}{64.7} 
    &\cellcolor{cyan!10}{87.8}  \\
    \midrule
     OpenAI-CLIP~\citep{radford2021clip} & L/14 &336 
    &76.6 &37.1/57.9
     &69.4
    &70.0
    &535
    &1591/351
    &73.3
    &40.8
    &76.9
    &64.5
    &87.6\\
    \rowcolor{cyan!10} \textsf{OpenVision} & \cellcolor{cyan!10}L/14 & \cellcolor{cyan!10}336 
   &\cellcolor{cyan!10}78.8 &\cellcolor{cyan!10}55.9/75.2
    &68.3
    &68.0
    &547
    &1520/310
    &73.3
    &45.3
    &75.4
    &64.4
    &88.1
\\
SigLIP~\citep{zhai2023sigmoid} &SoViT-400M/14 &384 &83.2 &52.0/70.2 &68.2
&61.3
&494
&1539/325
&72.9
&40.5
&74.7
&62.9
&86.8 
 \\
  \rowcolor{cyan!10}\textsf{OpenVision} & SoViT-400M/14 & 384 & 79.9  & 57.6/77.5
&67.4
&63.1
&540
&1500/353
&72.2
&43.7
&73.5
&63.4
&87.8

\\
    \bottomrule   
    \end{tabular}}
    \vspace{-.7em}
\end{table*}

\subsection{Visual Instruction Fine-tuning  and Evaluation}
\label{sec:llava_train_eval}
To assess the quality of visual encoders from the MLLM perspective, we benchmark them on general VQA tasks, which require generating free-form text answers based on visual inputs.  
Following prior practice~\cite{llava, llavanext, li2024llavaonevision},  we attach a lightweight MLP projector to the vision encoder, concatenate the resulting visual tokens to the language tokens, and perform visual instruction tuning. 
Unlike prior work that studies off-the-shelf checkpoints~\cite{tong2025cambrian}, we compare our \emph{from-scratch} \textsf{OpenVision} models with CLIP-style baselines at different sizes.
All experiments use Llama-3-8B as the language backbone and adopt two LLaVA setups:

\paragraph{1. LLaVA-1.5~\cite{liu2023improved}.} In this low-compute regime the vision encoder is kept frozen; only the lightweight projector and the language model are updated. This setup allows us to assess the quality of the pre-trained vision features. We train with the standard LCS-558K and LLaVA-665K datasets.

\paragraph{2. Open-LLaVA-Next~\cite{chen2024open}.} This high-compute regime gauges the encoder’s capacity for further learning and scaling. Roughly one million image–instruction pairs are used, and the vision backbone, projector, and LLM are \emph{all} fine-tuned. The setup also employs the ``any-resolution'' strategy \cite{llavanext} to tackle larger inputs: each image is resized to several aspect-ratio variants (\eg, $672\times672$, $336\times1344$) generated from a base size of $336\times336$.

\begin{table*}[t!]
    \centering
    \caption{Performance of \textsf{OpenVision} encoders at different scales with Llama3-8B under LLaVA-1.5.}
    \label{Tab:scaling}
    \vspace{-.6em}
    \resizebox{\linewidth}{!}{
    \begin{tabular}{c|c|c|c|c|c|c|c|c|c|c|c|c|c}
    \toprule   
    \multirow{2}{*}{\textbf{Vision Encoder}} &
    \multirow{2}{*}{\textbf{\# Res.}} &
    \multirow{2}{*}{\textbf{\# Params.}} &
    \multicolumn{2}{c|}{\textbf{CLIP-Bench}} &
    \multirow{2}{*}{\textbf{Text VQA}} &
    \multirow{2}{*}{\textbf{Chart QA}} &
    \multirow{2}{*}{\textbf{OCR.}} &
    \multirow{2}{*}{\textbf{MME}} &
    \multirow{2}{*}{\textbf{SEED}} &
    \multirow{2}{*}{\textbf{MMVet}} &
    \multirow{2}{*}{\textbf{SQA}} &
    \multirow{2}{*}{\textbf{GQA}}  &
    \multirow{2}{*}{\textbf{POPE}} \\ 
    \cline{4-5}
     & & & \textbf{Cls.} & \textbf{Retr.} & & & & & & & & & \\
    \midrule
    OpenAI-CLIP-L/14 & 224 & 303.7M  
    &75.5  &36.5/56.3 
    &56.1
    &13.2
    &177
    &1443/306
    &66.0
    &32.8
    &73.4
    &60.8
    &85.0\\
    L/14 &224  &303.7M &78.4  &55.3/75.2 
    &57.7
    &13.9
    &315
    &1487/317
    &69.5
    &35.2
    &73.6
    &62.9
    &86.4\\
  H/14  &224 &632.1M &80.4  &57.4/77.0
    &57.9 &13.6&330 &1501/308 &69.3&35.8 & 75.9&61.9&87.0 \\
\midrule
B/16 &224  &87.4M
&73.7
& 51.1/71.6
&54.1
&11.8
&262
&1496/293
&68.2
&30.9
&74.4
&61.6
&86.6\\
S/16  &224  &22.4M &65.9 &43.6/64.5 &51.8
&11.0
&202
&1348/264
&65.5
&24.6
&71.8
&60.1
&84.6
\\
Ti/16  &224 &5.9M &49.6 &50.0/30.4  &48.9
&11.7
&128
&1273/282
&59.9
&21.8
&71.8
&57.4
&82.0
\\
    \bottomrule
    \end{tabular}}
    \vspace{-.2em}
\end{table*}

\begin{table*}[t!]
    \centering
    \caption{Performance of \textsf{OpenVision} encoders with Qwen2.5-0.5B under LLaVA-1.5.}
    \label{Tab:small_llm}
    \vspace{-.6em}
    \resizebox{\linewidth}{!}{
    \begin{tabular}{c|c|c|c|c|c|c|c|c|c|c|c|c|c}
    \toprule   
    \multirow{2}{*}{\textbf{Vision Encoder}} &
    \multirow{2}{*}{\textbf{\# Res.}} &
    \multirow{2}{*}{\textbf{\# Params.}} &
    \multicolumn{2}{c|}{\textbf{CLIP-Bench}} &
    \multirow{2}{*}{\textbf{Text VQA}} &
    \multirow{2}{*}{\textbf{Chart QA}} &
    \multirow{2}{*}{\textbf{OCR}} &
    \multirow{2}{*}{\textbf{MME}} &
    \multirow{2}{*}{\textbf{SEED}} &
    \multirow{2}{*}{\textbf{MMVet}} &
    \multirow{2}{*}{\textbf{SQA}} &
    \multirow{2}{*}{\textbf{GQA}}  &
    \multirow{2}{*}{\textbf{POPE}} \\ 
    \cline{4-5}
     & & & \textbf{Cls.} & \textbf{Retr.} & & & & & & & & & \\
    \midrule
    {OpenAI-CLIP-B/16} & 224 &87.4M &68.3
    &33.1/52.4
      & 33.5 & 10.0 & 69 & 1059/255 & 49.1 & 13.6 & 55.8 & 48.5 & 82.3 \\ 
    {B/16} & 224 &87.4M    &73.9
    &51.1/71.6
    &34.8
    &10.1
    &132
    &1063/252
    &51.4
    &16.1
    &56.0
    &49.6
    &84.4\\
    \midrule
    {B/16} & 384   &87.4M
    &74.5
    &52.0/72.3
      & 38.2 &10.3 &174 &1171/280 &53.9 &15.9 &56.0 &51.8 &85.8 \\
    {S/16} & 384   &22.4M &67.1 &45.0/66.2
      & 32.8 & 9.9 & 78 & 1071/246 & 50.5 & 11.6 & 54.7 & 49.1 & 84.3 \\
    {Ti/16} & 384   &5.9M &51.4&32.2/53.0
      & 27.4 & 9.4 & 27 & 843/263 & 40.9 & 11.0 & 54.1 & 42.8 & 79.1 \\
    \bottomrule
    \end{tabular}}
\end{table*}

\paragraph{Evaluation benchmarks.} Performance is reported on a broad suite, including: MME~\cite{fu2023mme}, GQA~\cite{hudson2018gqa}, ChartQA~\cite{masry-etal-2022-chartqa}, POPE~\cite{Li-hallucination-2023}, TextVQA~\cite{singh2019towards}, OCR~\cite{ocr}, SEED~\cite{li2023seed}, MMVet~\cite{yu2024mm}, and SQA~\cite{lu2022learn}. 
We follow the \texttt{lmms-eval} protocol~\cite{zhang2024lmmsevalrealitycheckevaluation} for prompt formatting and use greedy decoding as the text generation strategy in all tasks.

\section{Main Results}
\label{sec:main results}
\subsection{\textsf{OpenVision} \vs Proprietary}
We compare our \textsf{OpenVision} family against popular proprietary and open-source vision encoders under the LLaVA-1.5 and Open-LLaVA-Next frameworks. To ensure fairness, all runs employ the original hyper-parameters provided by CLIPS~\cite{liu2024clips}, LLaVA-1.5~\cite{liu2023improved}, and Open-LLaVA-Next~\cite{chen2024open}.
Figure~\ref{fig:openvision_teaser} offers a high-level view: across nine representative benchmarks, \textsf{OpenVision} consistently matches---or surpasses---the performance of OpenAI's CLIP and Google's SigLIP.

A more comprehensive comparison is presented in Table~\ref{Tab:llava1.5_main1} and Table~\ref{Tab:llava-next_main1}, which also include results for LAION-2B-CLIP~\cite{schuhmann2022laion}, DataComp-1B-CLIP~\cite{gadre2023datacomp}, DFN-2B-CLIP~\cite{dfn}, and MetaCLIP-5B~\cite{metaclip}.
At $224\times224$ resolution, our \textsf{OpenVision-B/16} and \textsf{OpenVision-L/14} checkpoints significantly outperform their counterparts on most tasks under both MLLM setups. At $336\times336$ resolution, \textsf{OpenVision-L/14-336} either closely matches or exceeds OpenAI’s \mbox{CLIP-L/14-336} under Open-LLaVA-Next setup, establishing a new benchmark for open-source visual encoders.

These findings confirm that vision models trained entirely from public data and code can rival---and often outdo---proprietary alternatives, providing the research community with competitive, transparent, and flexible backbones for future multimodal work.

\subsection{More \textbf{\textsf{OpenVision}} Variants}
The full transparency of \textsf{OpenVision} allows us to freely craft a spectrum of vision encoders (see Appendix~\ref{app:vencoder} for architecture details) tailored to different resource or accuracy demands. Specifically, we illustrate this versatility by scaling \textsf{OpenVision} up/down and varying patch size for different application scenarios, and by showcasing its competitiveness even with an \textit{ultra-small} language model.

\begin{table*}[t!]
    \centering
    \caption{Impact of different patch sizes in LLaVA-1.5. Smaller patch sizes generally improve performance.}
    \label{Tab:patch_size}
    \vspace{-.6em}
    \resizebox{.9\linewidth}{!}{
    \begin{tabular}{c|c|c|c|c|c|c|c|c|c|c}
    \toprule
    
    \textbf{Vision Encoder} & \textbf{Patch Size} &
    \textbf{Text VQA} & \textbf{Chart QA} & \textbf{OCR.} &
    \textbf{MME} & \textbf{SEED} & \textbf{MMVet} & \textbf{SQA} &
    \textbf{GQA} & \textbf{POPE} \\
    
    \midrule
    Ti & 16 
    &50.2
&11.6
&139
&1329/280
&62.0
&21.4
&73.1
&58.0
&82.8\\
    \rowcolor{cyan!10}Ti & 8  &54.6
&12.9
&223
&1383/310
&66.3
&25.1
&73.1
&59.7
&85.3 \\
    \midrule
    S & 16 &54.3
&12.0
&235
&1393/343
&67.5
&28.8
&73.2
&61.6
&85.7
 \\
    \rowcolor{cyan!10}S & 8  &59.3
&15.9
&310
&1449/303
&70.3
&32.5
&74.7
&62.0
&87.1
\\
   
    \midrule
    B & 16 &57.9
&14.5
&293
&1432/333
&69.8
&33.2
&73.5
&62.8
&87.8
 \\
    \rowcolor{cyan!10}B & 8  &61.2
&17.2
&345
&1545/299
&71.8
&35.5
&74.0
&63.0
&87.0\\
    
    \bottomrule
    \end{tabular}}
    \vspace{-.3em}
\end{table*}

\begin{table*}[t!]
\centering
\caption{By pairing with a small LM (Smol-150M), we use \textsf{OpenVision-B/16-384} to create a $\sim$250M multimodal model. We show scaling behavior across Stage 2 data size, input resolution, and Stage 3 data size. 
We report performance on the following benchmarks: TextVQA, ChartQA, OCR-VQA, MME, SEED-Bench, MMVet, SQA, GQA, and POPE.}
\label{tab:scaling_all}
\vspace{-.6em}
\resizebox{\linewidth}{!}{
\begin{tabular}{l|l|l|ccccccccc}
\toprule
\textbf{Stage 2} 
& \textbf{Res.} 
& \textbf{Stage 3 Data Scale}
& \textbf{TextVQA} 
& \textbf{ChartQA} 
& \textbf{OCR-VQA} 
& \textbf{MME} 
& \textbf{SEED-Bench} 
& \textbf{MMVet} 
& \textbf{SQA} 
& \textbf{GQA} 
& \textbf{POPE} \\
\midrule
\multicolumn{12}{c}{\textit{(1) Scale Stage 2 Data: $\times$1, $\times$2, $\times$4, $\times$6, $\times$8 (fix resolution=384, Stage 3=LLaVA (665K))}} \\
\midrule
$\times$1 & 384 &\multirow{5}{*}{ LLaVA (665K)} &33.2
&10.3
&194
&743/212
&48.8
&15.8
&38.2
&54.2
&85.0
 \\
$\times$2 & 384 & &34.2
&10.6
&200
&785/204
&50.0
&16.4
&37.0
&54.3
&85.1
\\
$\times$4 & 384 & & 34.7
&10.2
&204
&760/210
&48.2
&16.3
&33.9
&54.4
&84.7 \\
$\times$6 & 384 & &34.7
&10.1
&223
&806/201
&47.4
&15.8
&37.5
&53.9
&84.6
 \\
$\times$8 & 384 & &35.4
&10.8
&234
&788/215
&45.1
&16.4
&35.6
&54.2
&84.7
 \\
\midrule
\multicolumn{12}{c}{\textit{(2) Scale Stage 3 Data: LLaVA (665K), LLaVA-Next (1M), LLaVA-One (3M) (fix Stage 2=$\times$8, Res=384)}} \\
\midrule
$\times$8 & 384 & LLaVA-Next (1M) &34.5
&26.1
&284
&869/219
&50.8
&16.4
&39.0
&53.9
&84.5
 \\
$\times$8 & 384 & LLaVA-OneVision (3M)  &36.3
&31.3
&319
&1051/248
&41.6
&20.7
&37.6
&53.3
&84.6
 \\
\midrule
\multicolumn{12}{c}{\textit{(3) Scale Input Resolution: 384→448→512→672→768 (fix Stage 2=$\times$8, Stage 3=LLaVA-OneVision (3M))}} \\
\midrule
$\times$8 & 448 & \multirow{4}{*}{LLaVA-OneVision (3M)} &37.0
&34.9
&333
&907/246
&41.3
&18.1
&36.8
&53.5
&85.0
\\
$\times$8 & 512 & & 38.2
&37.2
&347
&886/226
&39.3
&20.8
&39.0
&53.9
&86.0
 \\
$\times$8 & 672 & & 38.3
&43.2
&355
&1126/203
&46.6
&18.8
&43.7
&53.3
&85.5
 \\
$\times$8 & 768 & &40.6
&44.7
&382
&1080/242
&45.8
&22.0
&39.5
&53.2
&86.3
 \\

\bottomrule
\end{tabular}
}
\end{table*}

\paragraph{Scale Up for Superior Multimodal Performance.}
For applications demanding strong multimodal performance, larger vision encoders are beneficial as they can encode richer semantics and align more precisely with language. To this end, we release \textsf{OpenVision-H/14}, a 632.1 M-parameter vision encoder---significantly larger than the largest models from OpenAI’s CLIP and Google’s SigLIP.
As shown in Table~\ref{Tab:scaling} under the LLaVA-1.5 setup, this variant delivers substantial gains over OpenAI CLIP-L/14 in multimodal understanding, particularly in high-resolution VQA, OCR, and retrieval tasks, confirming the value of additional capacity for challenging multimodal tasks.

\paragraph{Scale Down for Resource-Limited Scenarios.}
To meet the memory and latency budgets of mobile or low-power devices, we train two compact variants, \ie, \textsf{OpenVision-S/16} and \textsf{OpenVision-Ti/16}.  In the same LLaVA-1.5 setting (Table \ref{Tab:scaling}), \textsf{S/16} retains 94\% of CLIP-L/14's average score while using more than $13\times$ fewer parameters, and \textsf{Ti/16} keeps 87\% at nearly $50\times$ smaller size.

We further pair these encoders with a 0.5 B-parameter Qwen-2.5 LLM \cite{qwen2.5}.  Firstly, simply replacing the baseline CLIP-B/16 with \textsf{OpenVision-B/16} boosts accuracy on nearly every benchmark (Table \ref{Tab:small_llm}). Then, by scaling down the size of vision encoder and meanwhile increasing the resolution from $224\times224$ to $384\times384$, the smaller \textsf{S/16} and \textsf{Ti/16} manage to maintain very competitive performance. These results confirm that lightweight, fully open vision backbones can power practical, high-quality edge-ready multimodal systems.

\paragraph{Variable Patch Sizes.}
In a ViT, the patch size determines the spatial resolution at which an image is tokenized \cite{wang2025scaling}, \ie, smaller patches supply finer details when encoding visual features (while at the cost of significantly increased computational budget).
To assess the impact of patch size, we therefore pre-trained two otherwise identical \textsf{OpenVision} models with $8\times8$ and $16\times16$ patches.

Table~\ref{Tab:patch_size} summarizes performance comparisons on a range of multimodal benchmarks under the LLaVA-1.5 setup. We can observe that the $8\times8$ variant delivers consistent and significant gains across all tasks, especially on fine-grained understanding tasks like TextVQA (\eg, +4.4\% for Tiny, +5.0\% for Small, and +3.3\% for Base).
However, we would also like to point out that these gains come at a cost: the finer patchification substantially increases the number of visual tokens, leading to much higher memory consumption and latency.

\subsection{\textbf{\textsf{OpenVision-Smol}}: Tuning with a 150M LM}
\label{Sec:smalls}
To push the portability of our vision backbones, we pair \textsf{OpenVision} with \emph{smol-LM}---a 150 M-parameter language model (LM), currently the smallest available on Hugging Face~\cite{allal2025smollm2}.  Specifically, we pair \textsf{OpenVision-B/16-384} with this Smol-150M, creating a multimodal system of fewer than 250M parameters---smaller than a ViT-L vision encoder on its own.

\paragraph{Three-stage training protocol.}
Following the training recipe of LLaVA-OneVision~\cite{li2024llavaonevision}, we first pre-train the models with image-caption alignment (Stage 1), and then perform additional vision-language pre-training using synthetic instructions (Stage 2); lastly, we fine-tune on curated multimodal instruction datasets (Stage 3).  
To probe scaling behavior, we systematically vary three knobs while holding all other hyper-parameters fixed: (1) the size of the Stage-2 instruction corpus, (2) the size of the Stage-3 instruction corpus, and (3) the input image resolution.

\begin{figure*}[t!]
    \centering
    \includegraphics[width=.95\linewidth]{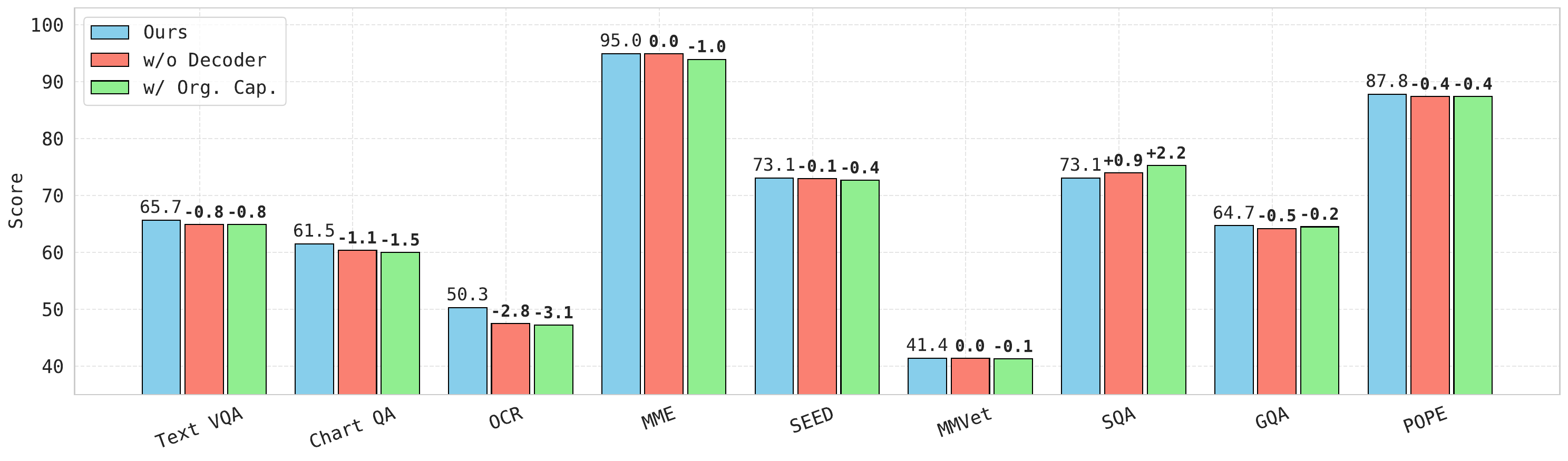}
    \vspace{-1.em}
    \caption{Ablations on the impact of an auxiliary decoder and synthetic captions. Results show that both contribute to better performance across multimodal benchmarks. We present performance gaps between different variants and our setting.}
    \label{fig:decoder_cap_ablation}
\end{figure*}

\begin{table*}[t!]
    \centering
    \caption{Ablation study on our \textsf{OpenVision} visual encoder with different input resolutions resulting from the three-stage training pipeline, evaluated under the Open-LLaVA-Next setting.
} 
    \label{Tab:resolution_next}
    \vspace{-.6em}
    \resizebox{.7\linewidth}{!}{
    \begin{tabular}{c|c|c|c|c|c|c|c|c|c}
    \toprule
     \textbf{Res.} & \textbf{Text VQA} & \textbf{Chart QA} & \textbf{OCR.} & \textbf{MME} & \textbf{SEED} & \textbf{MMVet} & \textbf{SQA} & \textbf{GQA} & \textbf{POPE} \\ 
    \midrule
    84$\times$84 &64.4  &63.1   &508  &1479/296   &71.5  &38.6  &72.5   & 63.5  &87.4   \\ 
   224$\times$224 &65.7
    &61.5
    &503
    &1567/332
    &73.1
    &41.4
    &73.1
    &64.7
    &87.8
 \\
     336$\times$336 &68.3
    &68.0
    &547
    &1520/310
    &73.3
    &45.3
    &75.4
    &64.4
    &88.1 \\   
    \bottomrule
    \end{tabular}}
\end{table*}

\paragraph{Main results.} Table~\ref{tab:scaling_all} reports the scaling results. Firstly, we can observe that enlarging the corpus in Stage 2 from $\times1$ to $\times8$ provides consistent gains on text-centric tasks such as TextVQA and OCR-Bench; although the gains flatten on reasoning-oriented suites like SEED-Bench and MMVet. Secondly, we notice that increasing data size in Stage 3 delivers a strong boost, especially in document-centric and chart reasoning tasks (\eg, ChartQA, OCR-Bench).  Lastly, raising the input resolution from $384$ px to $768$ px leads to the largest overall improvements, particularly for OCR and complex reasoning benchmarks.  

These results collectively confirm that our fully open \textsf{OpenVision} backbones retain strong scalability even when paired with a \underline{tiny} 150 M-parameter language model.  The resulting model family competitively offers a practical path to ultra-lightweight yet capable multimodal systems for real-world, resource-constrained deployments.

\section{Ablation Studies}
The results in Section \ref{sec:main results} show that \textsf{OpenVision} rivals, and sometimes surpasses, proprietary vision encoders such as OpenAI’s CLIP and Google’s SigLIP.  We now dissect the model to pinpoint the design choices that drive this performance.

\subsection{Auxiliary Decoder and Synthetic Caption}
Following CLIPS, \textsf{OpenVision} augments the standard contrastive objective with an auxiliary text decoder trained on the re-captioned Recap-DataComp-1B corpus.  Although CLIPS demonstrated that this generative signal improves cross-modal retrieval, its impact on \emph{multimodal reasoning} has not been examined.  We close this gap with two ablations: 1) \textbf{w/o Decoder}: remove the text decoder and train with pure contrastive loss; and 2) \textbf{w/ Orig.\ Caps}: keep the decoder but replace synthetic captions with the original DataComp-1B captions.

Figure~\ref{fig:decoder_cap_ablation} summarizes the findings.  We can observe that removing the decoder consistently degrades performance across most multimodal benchmarks, confirming that the generative objective supplies essential semantic supervision that the contrastive loss alone cannot provide. Additionally, replacing synthetic captions with the original, often noisy captions produces a similar drop, indicating that the richer, LLM-generated descriptions in Recap-DataComp-1B offer superior guidance for learning transferable visual features.

\paragraph{Takeaway.} With the results above, we can confirm that both components---the auxiliary text decoder and the high-quality synthetic captions---are critical to the strong multimodal performance of \textsf{OpenVision}.

\begin{table*}[t]
    \centering
    \caption{Ablation study on extending schedule higher-resolution fine-tuning in CLIPS pre-training as illustrated in Section~\ref{Sec:train_eval}. Doubling fine-tuning samples improves performance, especially in high-resolution tasks like OCR and ChartQA.}
    \label{Tab:extend}
    \vspace{-.6em}
    \resizebox{0.9\linewidth}{!}{
    \begin{tabular}{c|c|c|c|c|c|c|c|c|c|c}
    \toprule
     \textbf{224 $\times$ 224} &\textbf{336 $\times$ 336} & \textbf{Text VQA} & \textbf{Chart QA} & \textbf{OCR.} & \textbf{MME} & \textbf{SEED} & \textbf{MMVet} & \textbf{SQA} & \textbf{GQA} & \textbf{POPE} \\ 
    \midrule
 512M&    128M&68.6
&66.1
&513
&1574/326
&73.4
&40.7
&73.4
&65.0
&88.1\\
 1024M&   256M &68.3
&68.0
&547
&1520/310
&73.3
&45.3
&75.4
&64.4
&88.1\\
 512M&  512M &68.9
&68.6
&550
&1548/323
&74.0
&44.9
&73.9
&64.6
&88.3 \\
 0M&  768M
 &69.1
&68.2
&554
&1553/332
&74.2
&41.6
&71.9
&64.7
&88.5

  \\
    \bottomrule
    \end{tabular}}
\end{table*}

\subsection{Progressive Resolution Pre-training}
\label{Sec:progressive}
To significantly accelerate pre-training, \textsf{OpenVision} follows a three-stage curriculum that begins with very small crops and ends at $336/384$ px.  Prior works have shown that such schedules can accelerate CLIP training without hurting performance \cite{li2023clipa,li2023clipav2,li2023reclip,li2022flip}, but their downstream effect on multimodal performance---especially the contribution of the main low-resolution stages---has not been analyzed.

\begin{figure}[t!]
    \centering
    \includegraphics[width=\linewidth]{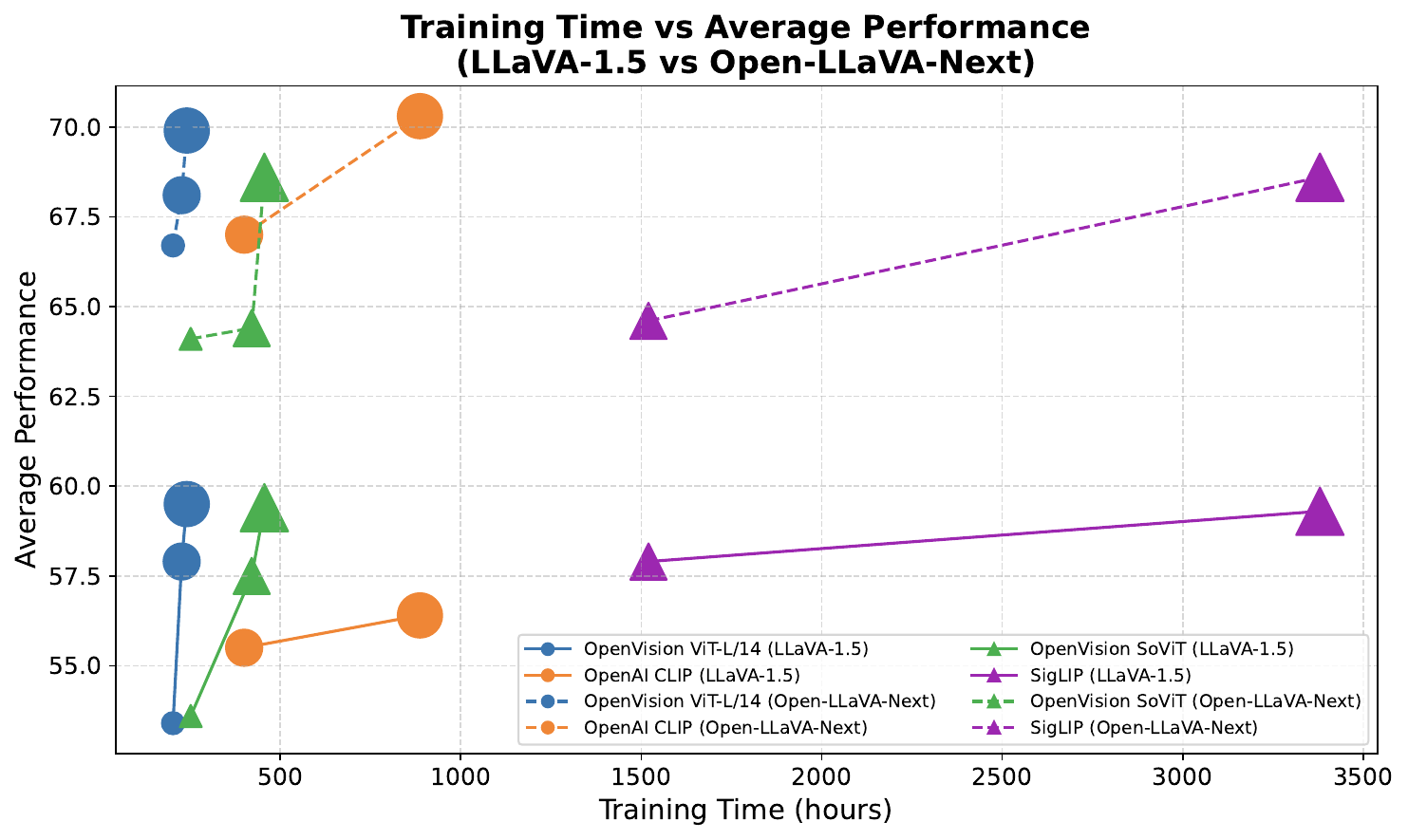}
    \vspace{-2em}
    \caption{Comparison of training time and average multimodal performance between our \textsf{OpenVision} and OpenAI-CLIP on both LLaVA-1.5 and LLaVA-Next. Larger markers correspond to vision encoders with higher input resolutions. As a fully open and cost-effective vision encoder, \textsf{OpenVision} achieves higher performance with significantly less pre-training time.}
    \vspace{-.5em}
    \label{fig:efficiency}
\end{figure}

To investigate this, we assess the performance of \textsf{OpenVision} encoders produced at the end of each stage using our multimodal evaluation pipeline (see Table~\ref{Tab:resolution_next} and more details in Appendix~\ref{app:resolution_viencoder}). Note that the LLaVA models built on these encoders use the same native resolution.  
Figure~\ref{fig:efficiency} also provides averaged multimodal performance against estimated training time and includes OpenAI's CLIP as a reference point.

We highlight two key findings from these results.  First, we can see that low-resolution pre-training is able to achieve competitive performance at a significantly reduced training cost. For example, the \textsf{OpenVision} encoder trained only at $84\times84$ resolution outperforms OpenAI's CLIP that is trained at $224\times224$ resolution in the Open-LLaVA-Next setting while requiring roughly only \textbf{half} of the pre-training compute.  
Second, training with progressively increasing resolutions (\textsf{OpenVision} $336\times336$) not only yields better performance than training at high resolution from scratch (OpenAI-CLIP $336\times336$) but is also $3\times$ more efficient in pre-training.

\paragraph{Takeaway.} These results confirm that progressive resolution training yields vision encoders that are both performant and computationally efficient for multimodal learning. 

\begin{table*}[h!]
    \centering
    \caption{Ablation study on the Stage 1 \& Stage 2 training data of small VLM. Results show that both contribute to better performance across multimodal benchmarks.} 
    \label{Tab:small_vlm_data}
    \vspace{-.6em}
    \resizebox{\linewidth}{!}{
    \begin{tabular}{c|c|c|c|c|c|c|c|c|c|c|c}
    \toprule
     \textbf{Stage 1} & \textbf{Stage 2} & \textbf{Stage 3} & \textbf{Text VQA} & \textbf{Chart QA} & \textbf{OCR.} & \textbf{MME} & \textbf{SEED} & \textbf{MMVet} & \textbf{SQA} & \textbf{GQA} & \textbf{POPE} \\ 
    \midrule
    LCS-558K &\XSolidBrush  &LLaVA-1.5  &19.5
&9.1
&92
&555/199
&25.1
&8.4
&35.5
&33.0
&61.8
  \\
     Recap-DataComp-558K&\XSolidBrush  &LLaVA-1.5  &20.7
&9.0
&59
&600/211
&24.1
&9.5
&35.0
&33.6
&68.9
 \\
     \midrule
     LCS-558K&OneVision-4M  &LLaVA-1.5  &19.2
&11.2
&191
&503/227
&24.0
&12.5
&34.8
&35.1
&65.0
\\
     LCS-558K&Recap-DataComp-4M  &LLaVA-1.5  &24.9
&10.6
&213
&720/210
&25.5
&15.3
&34.6
&37.6
&72.9 \\
    \rowcolor{cyan!10}Recap-DataComp-558K&Recap-DataComp-4M  &LLaVA-1.5  &26.5
&10.5
&136
&618/242
&26.2
&16.7
&36.5
&38.6
&72.7
\\
    \bottomrule
    \end{tabular}}
\end{table*}

\subsection{Extended High-Resolution Fine-Tuning}
The next interesting question we explore here is how much additional compute should be invested in the high-resolution stage. Using the number of image–text pairs processed as a proxy for training cost, our default CLIPS-style schedule fine-tunes on 512 M samples at $224\times224$, followed by 128 M samples at $336\times336$.  Doubling the budget, we compare three alternatives: (1) fine-tuning with 1024M samples at $224\times224$, followed by 256M samples at $336\times336$ (2) fine-tuning with 512 M samples at $224\times224$, followed by 512M samples at 336×336, and (3) fine-tuning entirely with 768M samples at $336\times336$. 

As reported in Table~\ref{Tab:extend}, all three strategies lead to consistent improvements over the baseline. The largest gains are observed in fine-grained tasks such as OCR and ChartQA, where high-resolution details are especially critical. Interestingly, while all extended training strategies yield improvements, diminishing returns emerge when training exclusively at $336\times336$. 

\paragraph{Takeaway.} These results suggest that a balanced allocation across resolutions is more efficient, as lower-resolution fine-tuning helps establish general visual representations while high-resolution tuning refines fine-grained understanding capabilities.

\subsection{\textbf{\textsf{OpenVision}} + \textbf{\texttt{Smol-LM}}}
Building on Section~\ref{Sec:smalls}, we further analyze a \emph{tiny} multimodal model that couples \textsf{OpenVision-B/16-384} with \texttt{Smol-LM} (150 M parameters).  
To deeper our understanding of its training dynamics, we hereby probe two factors: data source and learning rate.

Regarding data source, Table~\ref{Tab:small_vlm_data} demonstrates that increasing the amount of data consistently improves performance, regardless of whether the corpus is OneVision or Recap-DataComp.  
Data quality, however, has a stronger effect than quantity.  Substituting the 4 M-sample OneVision subset~\cite{li2024llavaonevision} with an equally sized slice of Recap-DataComp~\cite{li2024if} in Stage 2 yields substantial gains: \emph{TextVQA} improves from \textbf{19.2} to \textbf{24.9}, \emph{OCR-Bench} from \textbf{191} to \textbf{213}, and \emph{POPE} from \textbf{65.0} to \textbf{72.9}.  
Moreover, even a 558K-sample slice of Recap-DataComp outperforms the same-sized LCS baseline (\eg, \emph{TextVQA} \textbf{19.5} → \textbf{20.7}).
This suggests that Recap-DataComp not only scales better but is also a more effective source in multimodal learning.

For hyperparameter tuning, we present detailed results in Appendix~\ref{app:ablation_lr}. We can see that excessively low or higher learning rates degrade model accuracy, while an appropriately tuned learning rate is essential for maximizing performance, echoing the findings of~\cite{zhao2023tuning}.

\paragraph{Takeaway.}
In summary, our ablation studies emphasize that high-quality synthetic captions from Recap-DataComp and a moderate learning rate are critical is critical for maximizing the performance of this tiny multimodal models.

\section{Discussions}
From these experiments, we summarize three interesting observations on the vision encoder design when paired for multimodal learning:

\paragraph{1. Limited predictive value of CLIP benchmarks.} Traditional CLIP evaluation tasks---such as ImageNet~\cite{imagenet2009imagenet} classification accuracy and MSCOCO~\cite{lin2014coco} image-text retrieval---do not reliably predict a vision encoder’s performance in multimodal models. For instance, as shown in Tables~\ref{Tab:llava-next_main1} and Table~\ref{Tab:llava1.5_main1}, despite achieving a superior MSCOCO retrieval performance compared to OpenAI-CLIP, both LAION-2B-CLIP and DataComp-1B-CLIP do not exhibit corresponding advantages on multimodal benchmarks. Additionally, DFN-2B-CLIP, which attains state-of-the-art accuracy on ImageNet, similarly fails to translate this strength into improved multimodal task performance. These results suggest that strong image classification or retrieval metrics fail to capture the qualities needed for a vision encoder to be effective in multimodal foundation models.

\paragraph{2. Crucial role of generative training (auxiliary decoder).} The inclusion of an auxiliary text decoder with a generative loss (\eg, caption prediction) is essential for a vision encoder’s semantic understanding in multimodal models. To validate this observation, we conduct ablation experiments in Figure~\ref{fig:decoder_cap_ablation}, comparing the performance of vision encoder when trained with and without the auxiliary decoder. Results clearly demonstrate that removing the decoder significantly deteriorates the multimodal performance, indicating that generative training substantially enriches the encoder's learned visual representations beyond contrastive image-text learning alone. Specifically, the auxiliary decoder provides essential semantic supervision, allowing the encoder to acquire deeper visual insights beneficial for downstream multimodal reasoning tasks.

\paragraph{3. Benefits of training with synthetic captions.} Utilizing synthetic captions during pre-training is beneficial for enhancing the vision encoder's multimodal capabilities. We conduct ablation experiments in Figure~\ref{fig:decoder_cap_ablation} and demonstrate that replacing synthetic captions with original web-crawled captions results in a noticeable decline in multimodal performance, indicating that synthetic captions substantially enrich the learned visual representations beyond traditional web-crawled captions. Specifically, synthetic captions provide richer and more precise semantic supervision, enabling the vision encoder to achieve deeper visual understanding crucial for downstream multimodal reasoning tasks.

\section{Related Works}
\paragraph{Vision-Language Pre-training.} Vision-language pre-training serves as a foundational strategy for multimodal learning. The popular archiectures include ViLBERT~\cite{lu2019vilbert}, CLIP~\cite{radford2021clip}, and ALBEF~\cite{li2021align}, which employ independent encoders to separately process visual and textual inputs.
Recent advances in vision-language pre-training have been driven primarily by the development of innovative loss functions.
CoCa~\cite{yu2022coca} combines contrastive and generative training objectives within a unified encoder-decoder framework.
SigLIP~\cite{zhai2023sigmoid} further improves the original CLIP model by adopting a pairwise sigmoid loss. 
AIM-V2~\cite{fini2024multimodal} employs a multimodal autoregressive pre-training strategy, enabling large vision encoders to jointly model image and text tokens. 
CLOC~\cite{chen2024contrastive} strengthens localized vision-language alignment by introducing region-level contrastive learning.
Our work builds upon the recently proposed, fully-open CLIPS~\cite{liu2024clips} framework, which enhances CLIP by utilizing synthetic captions to enrich textual representations.

\paragraph{Open Vision Encoder for Multimodal Learning.}
Advanced closed-source multimodal models, such as OpenAI's GPT-4o  \cite{gpt4v,openai2024gpt4o}, Google's Gemini \cite{team2024gemini}, exhibit exceptionally strong vision language capabilities. However, because of their proprietary nature, the specifics of their visual processing mechanisms remain entirely unknown.
Recently open-source community make efforts to proposed fully-opened multimodal large language models which even achieve better performance like InternVL \cite{internvl}and LLaVA-OneVision~\cite{li2024llavaonevision}.
To develop high-performing MLLMs, the open-source community primarily focuses on curating high-quality, large-scale datasets, including vision-language alignment datasets~\cite{sharegpt4v,li2024if} and visual instruction datasets~\cite{llava,li2024llavaonevision,tong2025cambrian}. Meanwhile, others like ~\cite{internvl,nvlm2024} concentrate on novel architectural designs to better integrate state-of-the-art vision encoders with LLMs.
However, the selection of vision encoders is largely restricted to open-weight models such as CLIP~\cite{radford2021clip} and SigLIP~\cite{zhai2023sigmoid}. The challenge of training a fully open and high-performing visual encoder for MLLMs remains an open question.

\section{Conclusion}
This paper introduces \textsf{OpenVision}, a fully-open and cost-effective family of vision encoders designed to support the development of multimodal foundation models. Through extensive experiments, our \textsf{OpenVision} encoders demonstrate performance comparable to or surpassing widely used proprietary models like OpenAI's CLIP and Google's SigLIP. 
Furthermore, \textsf{OpenVision} scales flexibly in both model size and input resolution, making it suitable for deployment in diverse environments, ranging from large-scale computing infrastructures to edge devices. By releasing all model weights, code, and training data, we aim to foster research flexibility and drive further innovation in the community, paving the way for more transparent and adaptable multimodal foundation models.

{\small
\bibliographystyle{ieee_fullname}
\bibliography{egbib}
}

\newpage
\appendix
\section{Appendix}

\subsection{Ablation \wrt Input Resolutions of the Vision Encoder}
\label{app:resolution_viencoder}
Following Sec.~\ref{Sec:progressive}, we present model performance under the LLaVA 1.5 setting with varied input resolutions in Table~\ref{Tab:resolution_1.5}.
We draw a similar conclusion as the findings from the LLaVA-Next setting: a higher resolution into the vision encoder during training always help boost model performance on vision-language benchmarks.

\begin{table}[h!]
    \centering
    \caption{Ablation study on our \textsf{OpenVision} visual encoder with different input resolutions resulting from the three-stage training pipeline, evaluated under the LLaVA-1.5 setting.
} 
    \label{Tab:resolution_1.5}
    \vspace{-.8em}
    \resizebox{\linewidth}{!}{
    \begin{tabular}{c|c|c|c|c|c|c|c|c|c}
    \toprule
     \textbf{Res.} & \textbf{Text VQA} & \textbf{Chart QA} & \textbf{OCR.} & \textbf{MME} & \textbf{SEED} & \textbf{MMVet} & \textbf{SQA} & \textbf{GQA} & \textbf{POPE} \\ 
    \midrule
    84$\times$84 &50.4  &12.1   &231  &1372/290  &63.5  &28.8   &76.6   &58.8   & 83.9  \\ 
   224$\times$224 &57.7
    &13.9
    &315
    &1487/317
    &69.5
    &35.2
    &73.6
    &62.9
    &86.4   
     \\
     336$\times$336 &61.2
    &15.7
    &339
    &1525/315
    &70.5
    &36.2
    &75.1
    &63.7
    &87.2 \\
       
    \bottomrule
    \end{tabular}}
    \vspace{-1em}
\end{table}

\subsection{Visual Encoder Configuration} \label{app:vencoder}
We present detailed visual encoder configurations in Table~\ref{tab:visual-arch}. We demonstrate the flexibility of our approach by scaling \textsf{OpenVision} up/down and varying patch size for different application scenarios,  and by showcasing its adaptability even with very small language models. 

\begin{table}[h]
    \caption{\textbf{Visual encoder configurations} used in our paper.}
    \vspace{-.8em}
    \centering
    \resizebox{\linewidth}{!}{
    \begin{tabular}{l|c|ccc|c}
        \toprule
        Model Size & Patch Size & Layers & Width & Heads & \#Params (M) \\
        \midrule
        Tiny        & 16 or 8  & 12  & 192  & 3  & 5 \\
        Small           & 16 or 8  & 12  & 384  & 6  & 22 \\
        Base           & 16 or 8  & 12  & 768  & 12 & 86 \\
        Large           & 14  & 24  & 1024 & 16 & 303 \\
        SoViT-400M~\cite{alabdulmohsin2023sovit}          & 14      & 27  & 1152 & 16 & 412 \\
        Huge           & 14  & 32  & 1280 & 16 & 631 \\
        \bottomrule
    \end{tabular}
    }
    \label{tab:visual-arch}
    \vspace{-0.5em}
\end{table}

\subsection{Ablation \wrt Learning Rate}
\label{app:ablation_lr}
We also conduct comprehensive ablations \wrt the learning rate and other hyper-parameters during VLLM training. In Table~\ref{Tab:small_vlm_lr} shows that a mid-range learning rate setting of \(5{\times}10^{-5}\) (Stage 2 ViT) and \(5{\times}10^{-4}\) (Stage 3 LLM) achieves the best overall scores—\emph{TextVQA} 33.2, \emph{MME} 743/212, \emph{POPE} 85.0 --- whereas overly lower or higher rates degrade accuracy. Careful hyperparameter tuning is essential to maximize performance for practical and extensible multimodal pipelines.
\begin{table*}[t!]
    \centering
    \caption{Ablation study on the Stage 2 \& Stage 3's learning rate. Results show that both contribute to better performance across multimodal benchmarks.} 
    \vspace{-.8em}
    \label{Tab:small_vlm_lr}
    \resizebox{\linewidth}{!}{
    \begin{tabular}{c|c|c|c|c|c|c|c|c|c|c|c}
    \toprule
    \textbf{Stage 2} & \textbf{Stage 3 LLM} & \textbf{Stage 3 ViT} & \textbf{Text VQA} & \textbf{Chart QA} & \textbf{OCR} & \textbf{MME} & \textbf{SEED} & \textbf{MMVet} & \textbf{SQA} & \textbf{GQA} & \textbf{POPE} \\ 
    \midrule
    \rowcolor{cyan!10}1e-5   & 1e-5   & \multirow{11}{*}{\textit{\textbf{Frozen}}} &26.5
&10.5
&136
&618/242
&26.2
&16.7
&36.5
&38.6
&72.7

 \\ 
    1e-5   &\multirow{9}{*}{5e-4}   &  & 32.8  & 10.2  & 171   & 806/213 & 48.7  & 16.7  & 37.8  & 54.2  & 85.4 \\ 
    3e-5   &   &  & 33.2  & 10.6  & 173   & 759/215 & 47.8  & 17.0  & 38.1  & 54.9  & 84.7 \\ 
    \cellcolor{cyan!10}{5e-5}   &   &  & \cellcolor{cyan!10}33.2  & \cellcolor{cyan!10}10.3  & \cellcolor{cyan!10}194   & \cellcolor{cyan!10}743/212 & \cellcolor{cyan!10}48.8  & \cellcolor{cyan!10}15.8  & \cellcolor{cyan!10}38.2  & \cellcolor{cyan!10}54.2  & \cellcolor{cyan!10}85.0 \\ 
    7e-5   &   &  & 32.6  & 10.2  & 184   & 845/205 & 42.0  & 14.4  & 32.8  & 54.2  & 85.7 \\ 
    1e-4   &   &  & 32.5  & 9.4   & 165   & 734/211 & 48.1  & 14.4  & 37.8  & 53.1  & 85.4 \\ 
    3e-4   &   &  & 29.2
&9.2
&149
&649/205
&44.5
&14.1
&35.5
&50.5
&83.3\\ 
    5e-4   &   &  & 25.4  & 9.8   & 86    & 684/205 & 27.1  & 10.7  & 34.9  & 49.4  & 81.1 \\ 
    7e-4   &   &  &23.0
&9.2
&22
&812/210
&28.0
&14.4
&35.0
&47.5
&79.3\\ 
    1e-3   &   &  & 22.5  & 9.2   & 20    & 656/206 & 27.5  & 11.2  & 34.3  & 44.7  & 77.5 \\ 
    \midrule
     \multirow{9}{*}{5e-5}   & 1e-5  &\multirow{9}{*}{\textit{\textbf{Frozen}}} &26.1
&10.0
&147
&672/221
&24.8
&15.8
&34.1
&39.4
&78.2
 \\
        & 3e-5  & &29.1
&10.0
&178
&769/259
&26.9
&16.0
&35.6
&44.2
&80.7 \\ 
        & 5e-5  & &29.6
&10.0
&176
&797/240
&27.3
&15.8
&35.7
&46.3
&82.1 \\ 
        & 7e-5  & &30.4
&10.1
&185
&836/235
&27.2
&13.9
&35.3
&47.7
&83.3 \\ 
        & 1e-4  & &31.7
&9.8
&185
&876/260
&27.4
&13.9
&35.5
&9.4
&84.3 \\ 
        & 3e-4  & &32.8
&10.4
&198
&717/210
&44.5
&13.3
&36.9
&53.2
&84.7\\
        & \cellcolor{cyan!10}5e-4  &  & \cellcolor{cyan!10}33.2  & \cellcolor{cyan!10}10.3  & \cellcolor{cyan!10}194   & \cellcolor{cyan!10}743/212 & \cellcolor{cyan!10}48.8  & \cellcolor{cyan!10}15.8  & \cellcolor{cyan!10}38.2  & \cellcolor{cyan!10}54.2  & \cellcolor{cyan!10}85.0  \\
        & 7e-4  &  &32.4
&10.3
&191
&793/215
&49.5
&15.1
&35.9
&54.8
&86.3
 \\
        & 1e-3  & &32.1
&10.8
&202
&808/247
&50.2
&15.3
&31.6
&55.4
&85.5\\
\midrule
\multirow{2}{*}{5e-5}   &\multirow{2}{*}{5e-4} & 1e-6 &21.7
&9.3
&32
&705/223
&27.2
&12.5
&34.9
&46.2
&79.23

 \\
  &   &5e-6  &21.7
&9.3
&31
&706/223
&27.3
&12.8
&34.8
&46.1
&79.2
\\
    \bottomrule
    \end{tabular}}
    \vspace{-1em}
\end{table*}

\end{document}